\documentclass[letterpaper, 10 pt, conference]{ieeeconf}  % Comment this line out if you need a4paper
\IEEEoverridecommandlockouts                              % This command is only needed if 
\usepackage{hyperref}       % hyperlinks
\usepackage{url}            % simple URL typesetting
\usepackage{booktabs}       % professional-quality tables
\usepackage{amsfonts}       % blackboard math symbols
\usepackage{nicefrac}       % compact symbols for 1/2, etc.
\usepackage{microtype}      % microtypography
\usepackage[noadjust]{cite}
\usepackage{amsmath,amssymb}
\usepackage{gensymb}
\usepackage{algorithm}
\usepackage[algo2e]{algorithm2e}
\usepackage{color}
\usepackage{amsfonts}
\usepackage{graphicx}
\usepackage{tabulary,multirow,overpic,xcolor}
\usepackage{wrapfig}
\usepackage{sidecap}
\usepackage{wrapfig}
\usepackage{todonotes}
\usepackage{blindtext}
\usepackage{subcaption}

\hypersetup{
    colorlinks=true,
    linkcolor=blue,
    filecolor=magenta,      
    urlcolor=blue,
}

\newcommand{\app}{\raise.17ex\hbox{$\scriptstyle\sim$}}
\newcolumntype{x}[1]{>{\centering\arraybackslash}p{#1pt}}

\newlength\savewidth

\makeatletter\renewcommand\paragraph{\@startsection{paragraph}{4}{\z@}
  {.5em \@plus1ex \@minus.2ex}{-.5em}{\normalfont\normalsize\bfseries}}\makeatother

\title{\LARGE \bf 6-DOF Grasping for Target-driven Object Manipulation in Clutter}

\author{Adithyavairavan Murali$^{1,2}$, Arsalan Mousavian$^{1}$, Clemens Eppner$^{1}$, Chris Paxton$^{1}$, Dieter Fox$^{1,3}$% \\
\thanks{* This work is done while the first author was an intern at NVIDIA. $^{1}$NVIDIA {\tt \small \{amousavian, ceppner, cpaxton, dieterf\}@nvidia.com}, $^{2}$The Robotics Institute, CMU {\tt \small amurali@cs.cmu.edu}, $^{3}$University of Washington}
}

\begin{document}
\maketitle
%\newline \versionString

%===============================================================================
\begin{abstract}
% Robot grasping in clutter
% What's hard about the problem
% Our method
% results
%
Grasping in cluttered environments is a fundamental but challenging robotic skill.
It requires both reasoning about unseen object parts and potential collisions with the manipulator.
Most existing data-driven approaches avoid this problem by limiting themselves to top-down planar grasps which is insufficient for many real-world scenarios and greatly limits possible grasps. We present a method that plans 6-DOF grasps for any desired object in a cluttered scene from partial point cloud observations. Our method achieves a grasp success of 80.3\%, outperforming baseline approaches by 17.6\% and clearing 9 cluttered table scenes (which contain 23 unknown objects and 51 picks in total) on a real robotic platform. By using our learned collision checking module, we can even reason about effective grasp sequences to retrieve objects that are not immediately accessible. Supplementary video can be found \href{https://youtu.be/w0B5S-gCsJk}{here}.
\end{abstract}

%===============================================================================

\section{Introduction}
Grasping is a fundamental robotic task, but is challenging in practice due to imperfections in perception and control. Most commonly, grasp planning involves generating gripper pose configurations (3D position and orientation) that maximize a grasp quality metric on a target object in order to find a stable grasp. There are several factors that affect grasp stability, including object geometry, material, gripper contacts, surface friction, mass distribution, amongst several others~\cite{prattichizzo2008, bicchi2000robotic}. Most traditional approaches to grasping assume a separate perception system that can perfectly~\cite{prattichizzo2008}, or with some uncertainty~\cite{mahler2015gpgpisopt}, infer object information such as pose and shape. This is followed by physics-based grasp analysis~\cite{florianReview2013, prattichizzo2008} or nearest-neighbour lookup on a database of pre-computed grasps \cite{allengraspdatabase2009}. These methods are slow~\cite{goldfeder2011}, prone to perception error and do not generalize to novel objects.

\begin{figure}
    \centering
    \begin{tabular}{c@{\hskip3pt}c@{\hskip3pt}c}
         \includegraphics[width=0.15\textwidth]{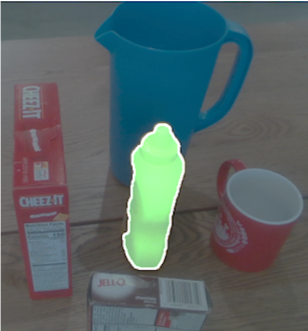}&
         \includegraphics[width=0.15\textwidth]{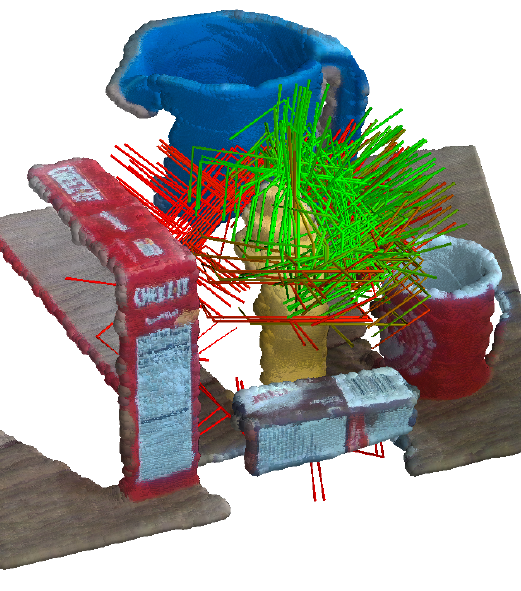}&
         \includegraphics[width=0.15\textwidth]{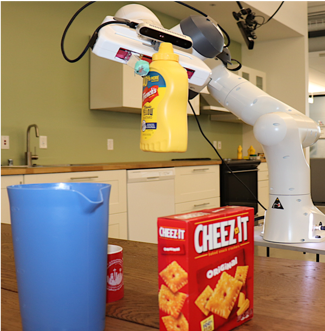}\\
         & 
    \end{tabular}
    \caption{Given an unknown target object~(\textit{left}) our proposed method leads to robust grasping~(\textit{right}) despite challenging clutter and occlusions.
    This is enabled by explicitly reasoning about successful and colliding grasps~(\textit{center}). \vspace{-4mm}}
    \label{fig:cover}
\end{figure}

% Paragraph Summary - Grasping in Clutter is hard
Grasp synthesis is much harder in clutter, such as the example in Fig~\ref{fig:cover}. The target object has to be grasped without any unwanted collisions with surrounding objects or the environment. In a real world application, a personal robot might be commanded to grasp a specific beverage from a narrow kitchen cabinet packed with other items. Grasps sampled agnostic of the clutter could end up in collision with the environment. Even if the gripper pre-shape is not in collision, it may be challenging to plan a collision-free and kinematically feasible path for the manipulator to achieve the gripper configuration. One would have to generate a diverse set of grasps since not all the grasps will be kinematically feasible to execute in the environment. Most model-based approaches in the grasping and task and motion planning literature assume perfect object knowledge or use an occupancy-grid representation for collision checking, which may not be reliable or practical in real-world settings~\cite{prattichizzo2008, kitaev2015physics, pushgrasping2010iros, berenson2007}.

% Paragraph Summary - grasping in clutter is hard due to perception
%% TODO --- CHRIS REMOVED THIS
%Perception becomes more challenging in clutter and this affects the quality of the grasp planner and collision checker. The point clouds are partially observable with potentially large and important parts of the object geometry being occluded by other objects.
%% CAN UNDO
A large part of the difficulty lies in perception. In clutter, large and important parts of object geometry are occluded by other objects. Traditional shape matching techniques will find it extremely challenging to operate in such conditions, even when object geometry is known. In addition, getting quality 3D information is challenging and previous methods resort to using high quality depth sensors \cite{mahler2017dex} or using multiple views \cite{ten2017grasp}, which would require observation-gathering exploratory movements impossible in confined spaces. This limits the deployment of such systems outside of controlled environments.
% A conservative heuristic such as grasping only the frontier objects, which are least occluded and planning a collision-free path is easier, would be extremely limiting in solving the task plan.

% \begin{figure}
%   \begin{center}
%     \includegraphics[width=1.0\linewidth]{figures/cover_v5.png}
%   \end{center}
%     % \missingfigure[figwidth=1.0\columnwidth]{Fancy Picture}%
%     \caption{Given the segmentation mask for the target object, our method generates grasps for the target object that are not colliding with the clutter.}
%     \label{fig:cover}
% \end{figure}

% Paragraph summary - Our learning-based work
Recent works have explored data-driven methods for grasping unknown objects \cite{ten2017grasp, bohg2014data, levine2016learning, mahler2017dex, pinto2016supersizing, robotsinhomenips2018, qtopt2018}. However, they mainly focus on the limited setting of planar-grasping and bin-picking. Some recent methods tackle the more difficult problem of generating grasps in $SE(3)$ from 2D (image) \cite{murali2017cassl}, 2.5D (depth, multi-view) \cite{ten2017grasp, lu2018planning, Yan-2018-113286} and 3D (point cloud)  \cite{6dofgraspnet, choi2018, PointNetGPD2019} data. These works primarily consider the problem from an object-centric perspective or in bin-picking settings. We consider the problem of 6-DOF grasp generation in structured clutter using a learning-based approach. Our method uses instance segmentation and point cloud observation from just a single view. We follow a cascaded approach to grasp generation in clutter, first reasoning about grasps at an object level and then checking the cluttered environment for collisions. We use a learned collision checker, which evaluates grasps for collisions from just raw partial point cloud observations and works under varying degrees of occlusion. Specifically, we present the following contributions:

\begin{itemize}
    \item A learning-based approach for  6-DOF grasp synthesis for novel objects in structured clutter, which uses a learned collision checker conditioned on the gripper information  and  on  the  raw  point  cloud  of the  scene.
    \item Showing that our approach, trained only with synthetic data, achieves a grasp accuracy of 80.3$\%$ with 23 real-world test objects in clutter. It also outperforms a clutter-agnostic baseline approach of 6-DOF GraspNet \cite{6dofgraspnet} with state-of-the-art instance segmentation \cite{chrisSegmentation} by 17.6\%.
    \item Demonstrating an application of our approach in moving blocking objects away out of the way to grasp a target object that is initially occluded and impossible to grasp.
\end{itemize}

\begin{figure*}[t]
\centering
    \includegraphics[width=0.99\linewidth]{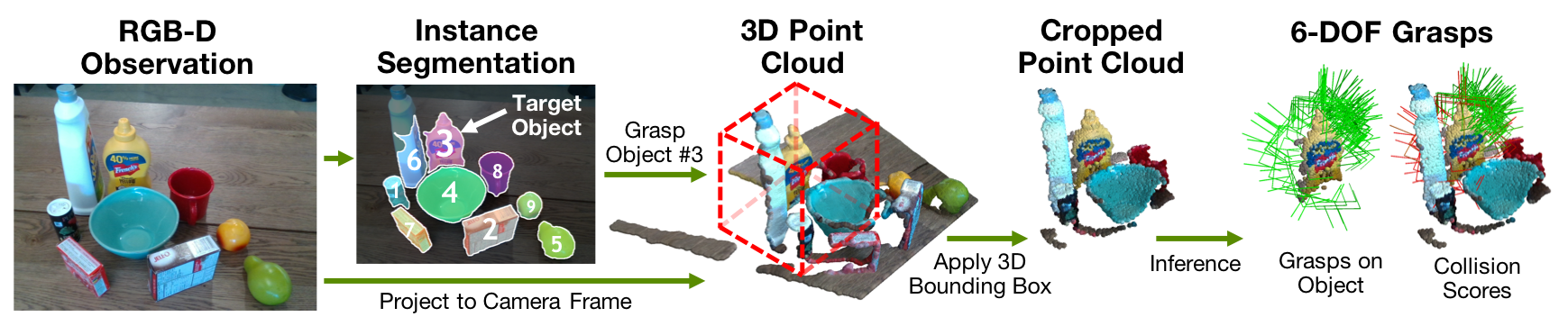}
  \caption{Overview of our cascaded grasping framework. A local point cloud centered on the target object is cropped from the scene point cloud using instance segmentation. 6-DOF grasps are then generated and ranked by collisions with the scene. \vspace{-4mm} }
\label{fig:system_overview}
\end{figure*}

\section{Related Work}
Grasping is a widely studied field in robotics~(\cite{prattichizzo2008, bicchi2000robotic, bohg2014data}).
In the following we will focus our comparison on existing approaches that are data-driven and the aspects in which they differ from the proposed method.
% Grasping is one of the fundamental problems in robotic manipulation and we refer readers to recent surveys~\cite{bicchi2000robotic, bohg2014data} for a comprehensive review. Classical approaches focused on physics-based analysis of stability~\cite{nguyen1988constructing} and usually require explicit 3D models of the objects. Recent papers have focused on data-driven approaches that directly learn a mapping from visual observations to grasp control~\cite{lenz2015deep, pinto2016supersizing, levine2016learning}. For large-scale data collection both simulation~\cite{mahler2017dex, mahler2016dex, 6dofgraspnet} and real-world robots~\cite{pinto2016supersizing, levine2016learning} have been used. \cite{mahler2017dex}~propose a versatile grasping model, that achieves $\>90\%$ grasping performance in the lab for the bin-picking task. 
% Talk about challenges in grasp sampling [Cite ISRR'19 paper, get full reference].

% First distinction: single objects vs. multiple ones
% example for single ones: \cite{lenz2015deep}, \cite{pinto15}, \cite{arsalan}, morris
% We: deal with multiple ones
\textbf{Grasping in clutter vs. isolated objects:}
Among learning-based methods for grasping a significant amount focuses on dealing with isolated objects on planar surfaces~(\cite{lenz2015deep,kappler2015leveraging,lu2018planning,yan2018learning,6dofgraspnet}).
Our approach specifically tackles the problem of grasping objects from a cluster of multiple objects.
This problem is significantly harder since the accessible workspace around the target object is severely limited, occlusions are more likely to hamper perception and predicting outcomes might be more difficult due to contact interactions between objects.
Although multiple learning-based approaches for dealing with grasping in clutter exist~(\cite{pinto2016supersizing,levine2016learning,mahler2016dexnet,ten2017grasp}) we will show in the following that they differ from our approach in multiple aspects.
% talk about non-learning based \cite{klingbeil2011grasping,dogar2011framework,herzog}

% Second distinction: Type of "clutter",  "bin picking" vs. "structured XXX clutter" (no bin to limit stuff)
% example for bin picking: mahler, handeye, 
% We: structured clutter, objects are in general bigger, toppling is considered more catastrophic
% From bin picking to structured clutter:  Collision avoidance becomes more important
\textbf{Bin-picking vs. structured clutter:}
Most learning-based grasping approaches for clutter deal with rather small and light objects that are spread in a bin~(\cite{pinto2016supersizing,levine2016learning,mahler2016dexnet,gualtieri2016high}).
% - small/homogeneous objects lead to more possible approach directions
In contrast our approach focuses on \textit{structured} clutter.
% TODO: revisit this definition
We define structured clutter as packed configurations of mostly larger, heavier objects.
Examples include kitchen cupboards or supermarket shelves.
Compared to the bin-picking setup successful grasps are more sparsely distributed in structured clutter scenarios.
Collisions and unintended contact is often more catastrophic since objects have fewer stable equilibria when they are not located on a pile.
% Also: Things can be retrieved without any pre-grasp manipulation
Since avoiding collision becomes more important, structured clutter is more prominent in evaluations of model-based task-and-motion-planning.
%TODO: ~(\cite{}).
Our approach explicitly predicts grasp configurations that are in collision and can do so despite occlusions.

% Another distinction: Planar vs Spatial Grasping
% the bin picking setup lends itself to a planar approach: mahler
\textbf{Planar vs. spatial grasping in clutter:}
Many learning-based grasp algorithms for clutter are limited to planar grasps, representing them as oriented rectangles or pixels in the image~(\cite{pinto2016supersizing,levine2016learning,mahler2018binpicking,zeng2018robotic}).
As a result, grasps lack diversity and picking up an object might be impossible given additional task or arm constraints.
This limitation is less problematic in bin-picking scenarios where objects are small and light.
% - bin-picking closely related to planar grasping (since the top-down view is the most informative one)
% - E.g. no upright standing bottles could be grasped
In structured clutter, spatial grasping is unavoidable, otherwise pre-grasp manipulations are needed~\cite{dogar2011framework}.
% - very few learning-based algorithms for spatial grasping exist
Those learning-based approaches that plan full grasp poses are either based on hand-crafted features~(\cite{klingbeil2011grasping,herzog2014learning,makhal2018grasping}) or have non-learned components~\cite{ten2017grasp}.
Our approach uses a learned grasp sampler that predicts the full 6D grasp pose and accounts for unseen parts due to occlusions.

% Third distinction: model-based methods vs. learning-based ones   (among approaches that grasp in structured clutter)
% examples for model-based approaches: dogar, cosgun,agboh,kitaev, plus all TAMP methods
% We: all models (collision, grasps) are learned
\textbf{Model-based vs. model-free:}
A lot of planning approaches exist that tackle scenarios of grasping in structured clutter~(\cite{cosgun2011push,dogar2011framework,king2015nonprehensile,kitaev2015physics,agboh2018real}).
% re-arrangement planning: \cite{king2015nonprehensile}
These approaches rely on full observability and prior object knowledge.
In contrast, our method does not require any object models and poses; grasps are planned based on raw depth images.
%  most structured clutter scenarios are tackled by planning algorithms that rely on full observability and/or prior object knowledge~\cite{}
In that regard, it is similar to other data-driven methods for clutter~(\cite{pinto2016supersizing,levine2016learning,ten2017grasp,mahler2018binpicking,zeng2018robotic}) but differs from techniques that use hand-engineered features~(\cite{klingbeil2011grasping,herzog2014learning,fischinger2015learning,makhal2018grasping}).

% Another distinction (?): semantics can be easily incorporated (because of the segmentation step)
% useful for high-level tasks such as search or manipulation
% examples that do not care about semantics: mahler
% those that do: mechanical search, semantic grasping
% We: consider segmentation and account for noisy segments in our models
% \textbf{Non-semantic vs. semantic grasping}
\textbf{Target-agnostic vs. target-driven:}
Few approaches focus on grasping specific objects in clutter~(\cite{jang2017end,danielczuk2019mechanical}).
Our method is target-driven as it uses instance segmentation~\cite{chrisSegmentation} to match grasps with objects.

\section{6-DOF Grasp Synthesis for Objects in Clutter}
\label{sec:approach}
% Start with assumptions
We consider the problem of generating 6-DOF grasps for unknown objects in clutter. The input to our approach is the depth image of the scene and a binary mask indicating the target object. In particular, we aim to estimate the posterior grasp distribution $P(G^{*}| X)$, where $X$ is the point cloud observation and $G^{*}$ is the space of successful grasps. We represent $g$~$\in G^{*}$ as the grasp pose $(R, T)$~$\in SE(3)$ of an opened parallel-yaw gripper that results in a robust grasp when closing its fingers.
% argument: why the distribution

% Talk about challenges of recovering the posterior distribution
The distribution of successful grasps is complex, multi-modal and discontinuous.
The number of modes for a new object is not known a-priori and is determined by the geometry, size, and physics of the object.
Additionally, small perturbations of a robust grasp could lead to failure due to collision or slippage from poor contact.
Finally, cluttered scenes limit the robot workspace significantly. Although a part of an object might be visible it could be impossible to grasp if the gripper itself is a large object~(such as the Franka Panda robot hand we use in our experiments) that leads to collisions with surrounding objects.

\subsection{Overview of Approach}
\label{subsec:overview}
% TODO: notation is not fully correct
Our cascaded grasp synthesis approach factors the estimation of $P(G^{*} | X)$ by separately learning the grasp distribution for a single, isolated object $P(G^{*} | X_{o})$ and a discriminative model $P(C | X, g)$ which we call \textit{CollisionNet} that captures collisions $C$ between gripper at pose $g$ and clutter observed as $X$. $X$ is the cropped point cloud of the scene and $X_{o} = \mathcal{M}_{o}(X)$ is the segmented object point cloud, where $\mathcal{M}_{o}$ is the instance mask of the target object.

The advantage of this factorization is twofold. First, it allows us to build upon prior work~\cite{6dofgraspnet} which successfully infers 6-DOF grasp poses for single, unknown objects. Second, by explicitly disentangling the reasons for grasp success, i.e., the geometry of the target object and a collision-free/reachable gripper pose, we can reason beyond simple pick operations. As shown in a qualitative experiment in Sec.~\ref{subsec:blocker} we can use our approach to infer which object to remove from a scene to maximize grasp success of the target object.

Fig.~\ref{fig:system_overview} shows an overview of our approach. During inference, a target object can be selected based on a state-of-the-art segmentation algorithm~\cite{chrisSegmentation}. Given this selection we infer possible successful grasps for the object ignoring clutter, and combine it with the collision results provided by CollisionNet.

In the following two sections, we will present both of these models. Note that our particular design decisions are based on comparisons with alternative formulations. In Sec.~\ref{subsec:ablation} we will show how our approach outperforms variants that do not distinguish between grasp failures due to collisions and target geometry, or use non-learned components.
% From P(G^{*} | X)& to $P(G_{i}^{*}| X_{i})$
% first evaluator: predicts the object-centric grasp robustness score agnostic of the clutter $P(S|X_{i}, g)$, with the training data being $G_{S}^{+} = G_{}^{+}$ and $G_{S}^{-} = G_{}^{-}$.
% second evaluator: $P(C|X_{i}, g)$

\subsection{6-DOF Grasp Synthesis for Isolated Objects}
\label{subsec:graspgeneration}
We first want to learn a generative model for the grasps given the point cloud observation of the cluttered scene. Though this generative model is learned from a reference set of positive grasps, it is not completely perfect due to several reasons. As a result, we follow the approach presented in \cite{6dofgraspnet} to have a second module to evaluate and further improve these generated grasps. Conditioned on the point cloud and grasp, the evaluator predicts a quality score for the grasp. This information could also be used to incrementally refine the grasp. We also explore the importance of object instance information in all stages of the 6-DOF grasping pipeline, from grasp generation to evaluation, in the ablation study.
% We refer readers to \cite{6dofgraspnet} for more details on the variational grasping approach.

\textbf{Variational Grasp Sampling:} The grasp sampler is a conditional Variational Autoencoder \cite{diederik2014} and is a deterministic function that predicts the grasp $g$ given a point cloud $X_{o}$ and a latent variable $z$. $P(z) = \mathcal{N}(0, I)$ is a known probability density function of the latent space. The likelihood of the grasps can be written as such: 

\begin{equation}
    P(G|X_{o}) = \int P(G|X_{o}, z) P(z) dz
\label{eq:generativemodel}
\end{equation}

Optimizing Eqn \ref{eq:generativemodel} is intractable as we need to integrate over all the values of the latent space \cite{diederik2014}. To make things tractable, an encoder $Q(z | X_{o}, g)$ is used to map each pair of point cloud $X_{o}$ and grasp $g$ to the latent space $z$ while the decoder reconstructs the grasp given the sampled $z$. The encoder and decoder are jointly trained to minimize the reconstruction loss $\mathcal{L}(\hat g, g)$ between the ground truth grasps $g \in G^{+}$ and predicted grasps $\hat g$, with the KL-divergence penalty between the distribution $Q$ and the normal distribution $\mathcal{N}(0, I)$: 

\begin{equation}
    \mathcal{L}_{VAE} = \displaystyle\sum_{z \sim Q, g \sim G^{*}} \mathcal{L}(\hat g, g) - \alpha \mathcal{D}_{KL} [Q(z|X_{o},g), \mathcal{N}(0, I)]
\label{eq:vae}
\end{equation}

Note that the input to the VAE is the point cloud of the target object segmented from the scene with instance mask. 

To combine the orientation and translation loss, we define the reconstruction loss as $\mathcal{L}(\hat g, g) = \dfrac{1}{n} \sum ||\mathcal{T}(g;p) - \mathcal{T}(\hat g;p)|| $ where $\mathcal{T}$  is the transformation of a set of predefined points $p$ on the robot gripper. During inference, the encoder Q is discarded and latent values are sampled from $\mathcal{N}(0,I)$. Both the encoder and decoder are based on the PointNet++ architecture \cite{pointnet}, where each point has a feature vector along with 3D coordinates. The features of each input point of the point cloud are concatenated to the grasp $g$ and the latent variable $z$ in the encoder and decoder respectively.
% The encoder learns to compress the relative information of point cloud X and latent variable grasp g in such a way that it can be reconstructed by the decoder. 

% TODO: To ablation studies
% Grasping can be formulated in object-agnostic and object-aware manner. In object-agnostic approach,  all the objects in the scene are considered and the model needs to generate grasps for each of them. However, the problem becomes quite challenging as the space of solutions grow  exponentially with the increase in number of  objects which makes the training intractable. Instead, we use the object instance information to constrain the complexity of the space of grasps and steer the generated grasps toward the target object.

Though instance information can give a strong prior about the object, it is not perfect in practice. This is especially the case in cluttered scenarios where objects are occluded or very close to each other, resulting in noisy under and over-segmentation. When rendering the segmentation in simulation, we add random salt-and-pepper noise to the object boundaries and randomly merge partially occluded objects to neighboring ones in image space, to mimic the imperfections of instance segmentation methods on the real images.

\textbf{Grasp Evaluation:}
Though the grasp sampler is trained with only positive grasps, it may still predict failed grasps which need to be identified and removed.
% Conditioned on the grasp and point cloud the evaluator predicts a quality metric $P(Q|X, g)$.
We train an evaluator that predicts a grasp score $P(S|X_{o}, g)$, with the training data consisting of positive~$G_{S}^{+} = G_{}^{+}$ and negative~$G_{S}^{-} = G_{}^{-}$ grasps. The evaluator's input is $X_{o}$, the point cloud of the object segmented from the full scene.
Since the space of all possible 6-DOF grasp poses is large, it is not possible to sample all the negative grasps for training the grasp evaluator~$P(S|X_{o}, g)$.
Therefore, during training we sample from true negatives but also sample hard negative grasps by perturbing positive grasps with a small translation and orientation and choosing those that are in collision with the object or are too far from the object to grasp the object.
At test time on the robot, the grasps are ranked by their evaluator scores and only those above a threshold are selected.
% as follows:

% \begin{equation}
%     \mathcal{L}_{evaluator} = -(ylog(s) + (1-y)log(1-s))
% \end{equation}

\textbf{Grasp Refinement:}
A significant proportion of the grasps rejected by the evaluator are actually in close proximity to robust grasps.
This insight could be exploited to perform a local search in the region of $g$ to iteratively improve the evaluator score.
We concretely want to sample $\Delta g$ to increase the probability of success, i.e., $P(S|\Delta g+g,X_{o}) > P(S|g,X_{o})$. 
The refinement was found using gradient descent in~\cite{6dofgraspnet}.
In practice, computing gradients is not fast. Instead, we use Metropolis-Hastings sampling where a random~$\Delta g$ is sampled and with probability of $\frac{P(S|g+\Delta g, X_{o})}{P(S|g, x)}$ grasp $g + \Delta g$ is accepted.
We observe that this sampling scheme yields similar performance to the gradient-based one while it is computationally twice as fast.

% talk about differences w.r.t. graspnet

\subsection{Collision Detection for Grasps in Clutter: CollisionNet}
\label{subsec:collision detection}
CollisionNet predicts a clutter-centric collision score $P(C| X, g)$ given the full scene information $X$.
% The training data for CollisionNet is $G_{C}^{+} = \{g| g\in G^{+} \cup G^{-}, \neg\Psi(g, \textbf{x}) \}$ and $G_{C}^{-} = \{g| g\in G^{+} \cup G^{-}, \Psi(g, \textbf{x}) \}$.
The training data for CollisionNet is $G_{C}^{+} = \{g| g\in G_{ref}, \neg\Psi(g, \textbf{x}) \}$ and $G_{C}^{-} = \{g| g\in G_{ref}, \Psi(g, \textbf{x}) \}$.
The ground truth labels are generated in simulation with a collision checker $\Psi$ assuming full state information $\textbf{x}$. In each batch, we used balanced sampling of grasps within the subsets of the reference set $G_{ref}$, which consists of the positive and negative sets ($G_{}^{+}$, $G_{}^{-}$), hard-negatives generated by perturbing positive grasps ($G_{hn}^{-}$) and grasps in free space $G_{free}$. We observed that balanced sampling improved the stability of training and generalization at test time over uniform sampling from $G_{}^{+} \cup G_{}^{-}$. Similar to the grasp evaluator, the scene/object point cloud $X$/$X_{o}$ and gripper point cloud $X_{g}$ are combined into a single point cloud by using an extra indicator feature that denotes whether a point belongs to the object or to the gripper.
The PointNet++~\cite{pointnet} architecture then uses the relative information between gripper pose $g$ and point clouds for classifying the grasps. CollisionNet is optimized using cross entropy loss. 

% \add{[Not sure where's the best place to put this, alternative is Section \ref{subsec:realrobot}. CollisionNet only considers collisions between the gripper and the clutter. In real robot experiments, we also use occupancy voxel-grid collision checking (described in \ref{subsec:realrobot}) on top of CollisionNet to make sure that the rest of the manipulator arm does not collide with the clutter and table during motion planning.]}\arsalan{put it in \ref{subsec:realrobot}}

\subsection{Implementation Details} 
Training data is generated online by arranging multiple objects randomly at their stable poses. Objects are added to the scene with rejection sampling poses to ensure they are not colliding with existing clutter. In order to generate grasps for the scenes, we combine the positive and negative grasps of each object from \cite{6dofgraspnet} which includes a total of 126 objects from several categories (boxes, cylinders, bowls, bottles, etc.). From each scene we take multiple 3D crops centered on the object (with some noise) along with grasps that are inside the crop. The cropped point cloud of the 3D box is down-sampled to $4096$ points. All the samplers and VAEs are based on PointNet++ \cite{pointnet} architecture and the dimension of latent space is set to 2. During inference, object instances are segmented with \cite{chrisSegmentation}. The VAE sampler generates the grasps given the point cloud of the target object by sampling 200 latent values. Grasps are further refined with 20 iterations of Metropolis-Hastings. The whole inference takes 2.5s on a desktop with NVIDIA Titan XP GPU.

%The filtered grasps are then ranked by \textit{CollisionNet}, the second evaluator which predicts a clutter-centric collision score $P(C| X, g)$ given the full scene information $X$.
% There are multiple ways in which we can use the object instance information.
% %to disambiguate the scene and provide prior knowledge about the object. 
% One way is to provide it as an additional feature vector to the clutter-centric sampler and evaluator (as shown in Fig \ref{fig:segmodes}b), by concatenating a indicator mask $\mathcal{M}_{i}$ to the features of each point in the input point cloud $X$. An alternative would be to use the instance mask to get a segmented point cloud of the object $X_{i} = \mathcal{M}_{i}(X)$ and perform object-centric grasp sampling, followed by clutter-centric evaluation (as shown in Fig \ref{fig:segmodes}c), the details of which will be explained in \ref{sec:graspgen}.

% \subsection{Collision Checking at Test Time}
% CollisionNet is awesome.

% \begin{itemize}
%     \item Dataset Generation
%         \begin{itemize}
%         \item Grasps - object centric
%         \item Clutter - Collisions and Occlusions
%         \end{itemize}
%     \item Variational Grasp Generation
%         \begin{itemize}
%             \item Sampling, evaluation, refinement
%         \end{itemize}{}
%     \item Object-Instance information
%     \item heuristic for Task Planning

% \end{itemize}
% \begin{align}
%     \alpha_{i} = \left|\frac{\partial P(c|X_{i}, g)}{\partial X_{i}}\right|
% \end{align}

\section{Experimental Evaluation}

%We used FleX~\cite{flex2014} physics engine to generate ground truth reference sets of successful and unsuccessful grasps. See \cite{6dofgraspnet} for details of grasp generation for individual objects. 
%In total, we use ? objects from five categories in ShapeNet \cite{shapenet} including bowls, bottles and some randomly generated boxes and cylinders. Data was rendered in an online fashion. \adithya{TODO: Explain more}

%\textbf{Network Architecture:} All the samplers and evaluators use the PointNet++ architecture. We use latent size of 2 for all samplers. \adithya{TODO: Explain}

\subsection{Ablation analysis and Discussion}
\label{subsec:ablation}
\textbf{Evaluation Metrics:} Following \cite{6dofgraspnet}, we used two metrics for evaluating the generated grasps: success rate and coverage. Success rate is the percentage of grasps that succeed grasping the object without colliding and coverage is the percentage of sampled ground truth grasps that are within $2cm$ of any of the generated grasps. The ablations were done in simulation with a held-out test set of 15 unseen objects of the same categories arranged at random stable poses in 100 different scenes. Physical interactions are simulated using FleX~\cite{flex2014}. Area under curve (AUC) of the success-coverage plot is used to compare different variation of the methods in the ablations.

%The following experiments quantitatively evaluate the proficiency of the models in generating grasps that are simultaneously robust for grasping a target object while being collision free considering the cluttered scene. Once grasps were sampled from the synthetic point clouds, they were physically executed in Flex \cite{flex2014} to determine their grasp \textit{Success Rate}. Grasps that were in collision with the scene objects are considered a failure, regardless of their robustness in grasping the object. We also measure the diversity of the grasps generated with a \textit{Coverage} metric (the x-axis in Fig \ref{fig:cascaded_single_stage}), which is computed by comparing the predicted set of grasps $\hat G$ with the reference set of collision free positive grasps $\{g: g \in G_{}^{+}, \neg\Psi(g, \textbf{x})\}$. The Area Under Curve (AUC) is also computed from the Success-Coverage curve.

\begin{figure}
  \begin{center}
    \includegraphics[width = 0.7\linewidth]{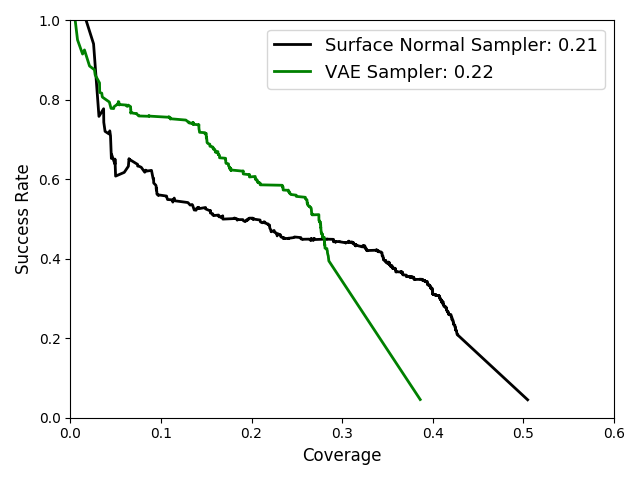}
  \end{center}
 \vspace{-2mm}
  \caption{Comparing the VAE sampler and Surface Normal Sampler. The number next to the legend is the area under curve (AUC) and the VAE sampler has a higher AUC.}
  \vspace{-2mm}
\label{fig:curve_geometry}
\end{figure}

\begin{figure}
  \begin{center}
    \includegraphics[width = 0.7\linewidth]{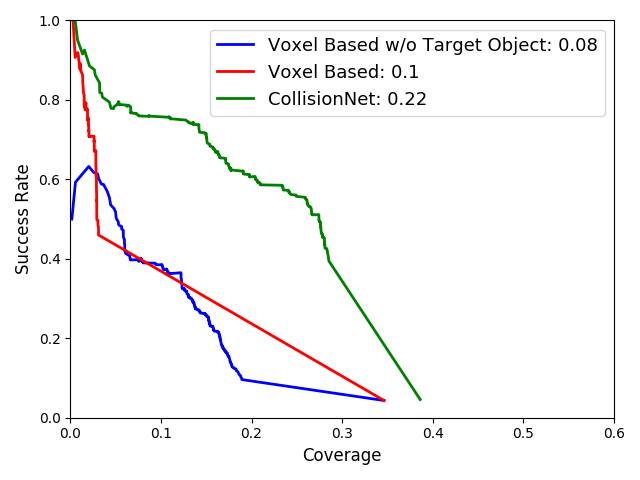}
  \end{center}
\vspace{-2mm}
  \caption{CollisionNet outperforms the Voxel-based approach in both success and coverage. The Voxel-based without Target Object ablation only considers collisions with the scene.}
\vspace{-6mm}
\label{fig:curve_voxel}
\end{figure}

\textbf{Learned vs. Surface Normal Based Grasp Sampler:} The first ablation study we consider is the effect of using a learned VAE to sample grasps in comparison with a geometric baseline. This baseline generates grasps by sampling random points on the object along surface normals, with random standoff, and random orientation along the surface normal. Fig.~\ref{fig:curve_geometry} shows that our learned VAE sampler yields more grasp coverage. It is worth noting that the surface-normal based sampler performed well for simpler shapes like boxes but failed to generate grasps for more complex geometry with rim, handles, etc.

%We also benchmarked the VAE sampler with a non-learning geometric baseline, specifically the algorithm used to generate the training data. The grasps are sampled by first estimating the surface normal from the point cloud and applying a random standoff and planar rotation to get the grasp. Once the grasps are sampled, they are evaluated and refined in the same way. As shown in Fig \ref{fig:curve_geometry}, our variational sampling approach in clutter outperfomed the geometric sampler. Though the geometric heuristic works decently for simple shapes like boxes, it fails to generate grasps for complex geometry with rims, handles, etc.

\begin{figure}
  \begin{center}
    \includegraphics[width = 0.99\linewidth]{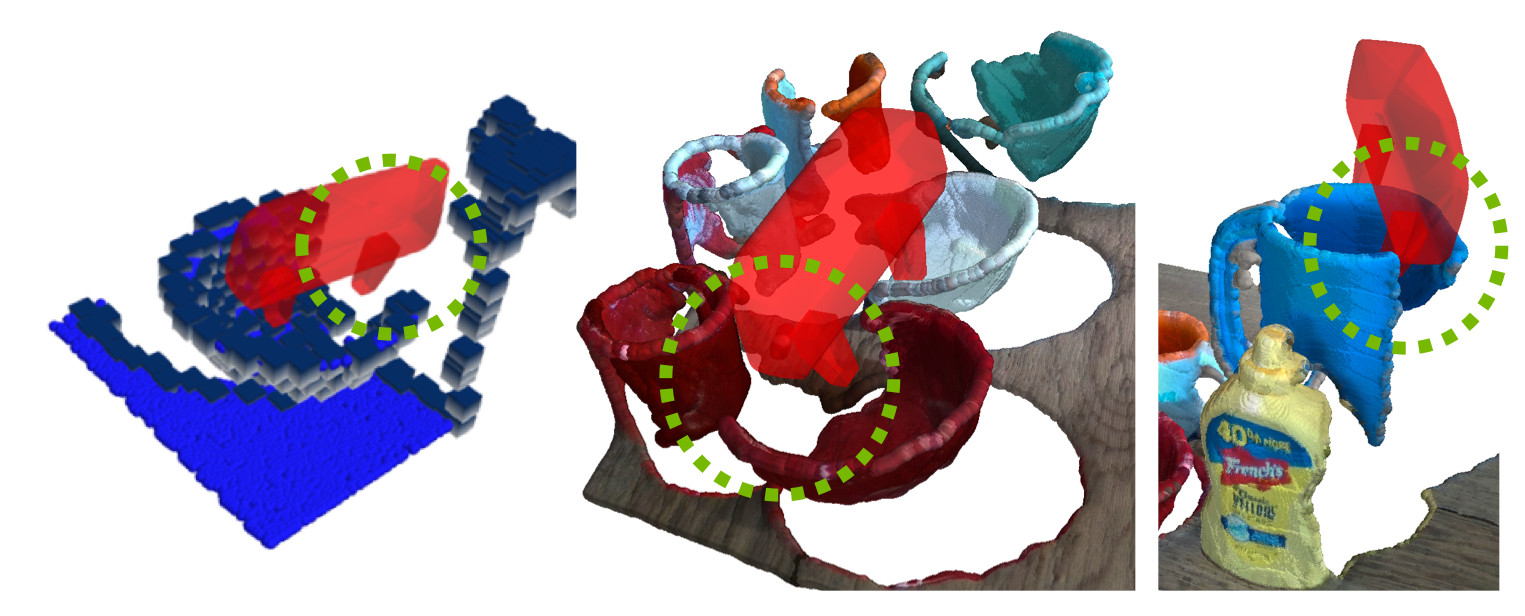}
  \end{center}
  \caption{Examples where the voxel-based heuristic fails to predict collisions but CollisionNet succeeds. These false positives are due to missing points (region highlighted by dotted circle) from occlusion. These grasps will lead to critical collisions if executed. \vspace{-4mm}}
\label{fig:voxel_failure_mode}
\end{figure}

\textbf{CollisionNet vs Voxel-Based Collision Checking:} We compared the effectiveness of CollisionNet with a voxel-based heuristic commonly used (such as in MoveIt!~\cite{chitta2012moveit}) for obstacle avoidance in unknown 3D environments. In our case, from each object, 100 points are sampled using farthest point sampling. Each sampled point is represented by a voxel cube of size $2cm$. Collision checking is done by checking the intersection of the gripper mesh with any of the voxels. As shown in Fig.~\ref{fig:curve_voxel}, CollisionNet outperforms the voxel-based heuristic in terms of precision and coverage. Qualitatively, we observed that the voxel-based representation fails to capture collision when the gripper mesh intersects with occluded parts of objects, or if there is missing depth information (see Fig.~\ref{fig:voxel_failure_mode}). In cases where the voxel-based collision checking fails, CollisionNet has $89.7\%$ accuracy in classifying the collisions correctly.

The voxel-based approach also has several false negatives by rejecting good grasps that are slightly penetrating voxels corresponding to points on the target object, as 
%The voxels are more conservative and
the voxels expand the spatial region for collision checking. Without considering the voxels on the target object for collisions, the coverage decreases marginally (blue curve in Fig.~\ref{fig:curve_voxel}). The grasp success also decreases as grasps that are actually colliding with the target object are not pruned out. CollisionNet does not suffer from such biases and can reason about relative spatial information in the partial point clouds.
% \begin{figure}
%   \begin{center}
%     \includegraphics[width = 0.8\linewidth]{figures/voxelnet-collisionnet.png}
%   \end{center}
%   \caption{Examples of incorrect predictions by the voxel heuristic where CollisionNet succeeds. Left: False Positive due to occlusion Right: False Negative due to discretization error when a \adithya{TODO: Replace with real pc examples}}
% \label{fig:voxel_failure_mode}
% \end{figure}

%\arsalan{Remove and move to supplemantary video: In Fig \ref{fig:voxel_failure_mode}(a), the target object (the bowl) is missing some points due to partial occlusion (i.e. no collision objects are added in that region) and the grasp penetration into the target object goes undetected. Since CollisionNet is able to reason about the relative spatial information in the point set, it is able to generalize to even occluded parts of the scene point cloud. In Fig \ref{fig:voxel_failure_mode}(b), the heuristic rejects grasps that are slightly penetrating the collision cubes, which cover a larger region than the object itself. On the test set, CollisionNet is able to successfully predict collisions $89.7\%$ of the time in cases where the heuristic fails. Overall, CollisionNet improves the quality and resolution of collision checking needed for target-driven 6-DOF grasping in clutter. }

\begin{figure}
  \begin{center}
    \includegraphics[width = 0.7\linewidth]{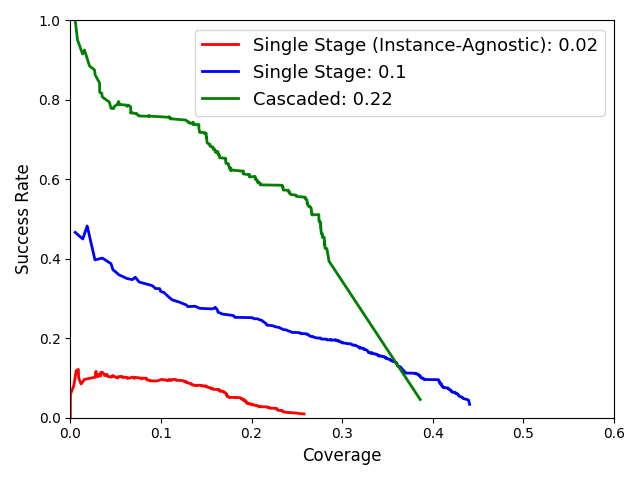}
  \end{center}
  \caption{Our cascaded approach demonstrates much higher success and coverage compared to a single-stage and instance-agnostic model.}
\label{fig:cascaded_single_stage}
\end{figure}

\begin{figure}
  \begin{center}
    \includegraphics[width = 0.99\linewidth]{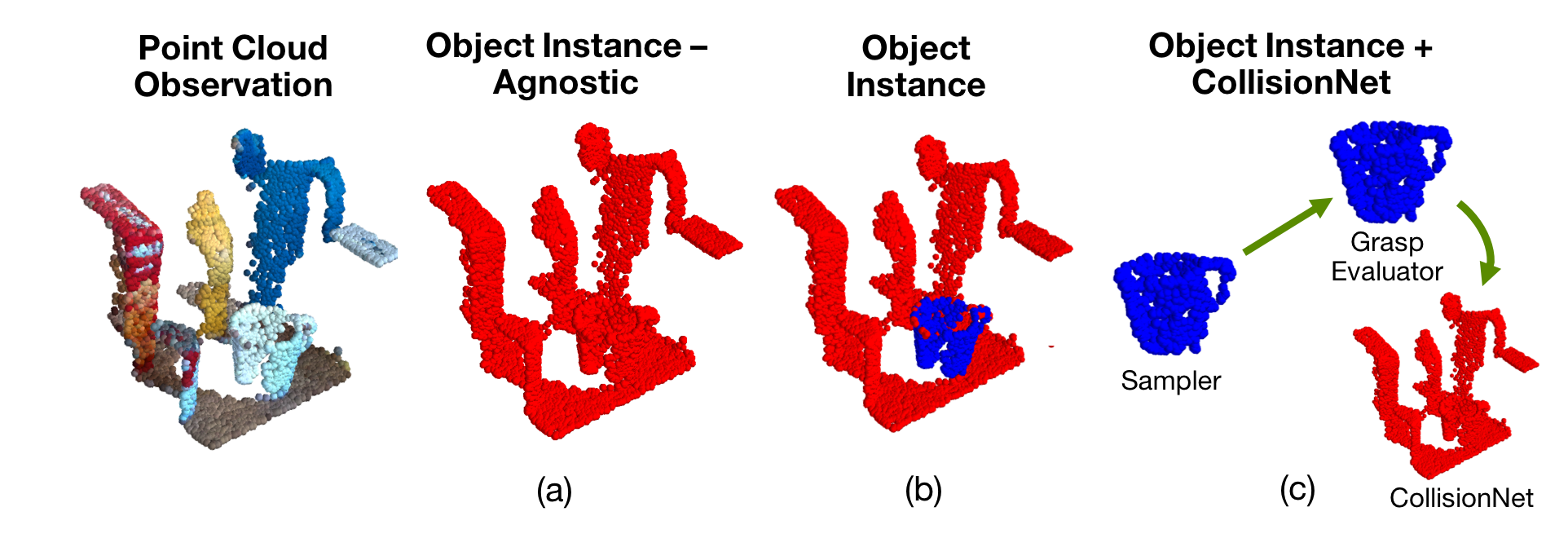}
  \end{center}
  \caption{Representative example of the data provided to the different grasping architectures a) single-stage model without object instance information b) single-stage model with object instance mask used as a feature vector along with the point cloud c) our cascaded model to sample with object-centric point cloud and evaluate for collisions with clutter-centric data. The target object is colored in blue. \vspace{-5mm}}
\label{fig:segmodes}
\end{figure}

\textbf{Single-stage vs. Cascaded Evaluator:} Instead of a cascaded grasp generation approach, one could also use a \textit{single-stage} sampler and evaluator with object instance information.  Once the grasps are sampled, there is only a single evaluator that directly estimates $P(S, \lnot C | X, g)$.
%$Q$ is the score of successful grasps for the target object that do not collide with anything in the scene (including the target object).
The positive training set is $G_{SC}^{+} = \{g| g\in G^{+}, \neg\Psi(g, \textbf{x}) \}$ while the negative set is $G_{SC}^{-} = \{g| g\in G^{+}, \Psi(g, \textbf{x}) \} \cup \{g|g\in G^{-} \}$. As a result, some positive grasps $g \in G^{+}$ will be in collision resulting in lower scores. An example of the input data to this baseline is shown in Fig.~\ref{fig:segmodes}(b), where the indicator mask of the target object is passed as an additional feature to the PointNet++ architecture. We found that the cascaded model outperformed the single-stage model, as shown in Fig \ref{fig:cascaded_single_stage}.

%This is due to several reasons.
This improvement is due to two factors. First, the VAE is far more proficient in learning grasps from an object-centric observation than from scene-level information. Second, the cascaded architecture imposes an abstraction between having a grasp evaluator that is singly proficient in reasoning about grasp robustness and CollisionNet that is proficient in predicting collisions. %This approach is shown to be superior to a monolithic model that jointly estimates grasp robustness and collision avoidance, both of which are extremely complex tasks in structured clutter\chris{~\cite{}}.

% CE: This should go to the ablation studies
% Use this figure to explain ablations
% It also needs a new caption

\begin{figure*}[t]
\centering
    \includegraphics[width=0.99\linewidth]{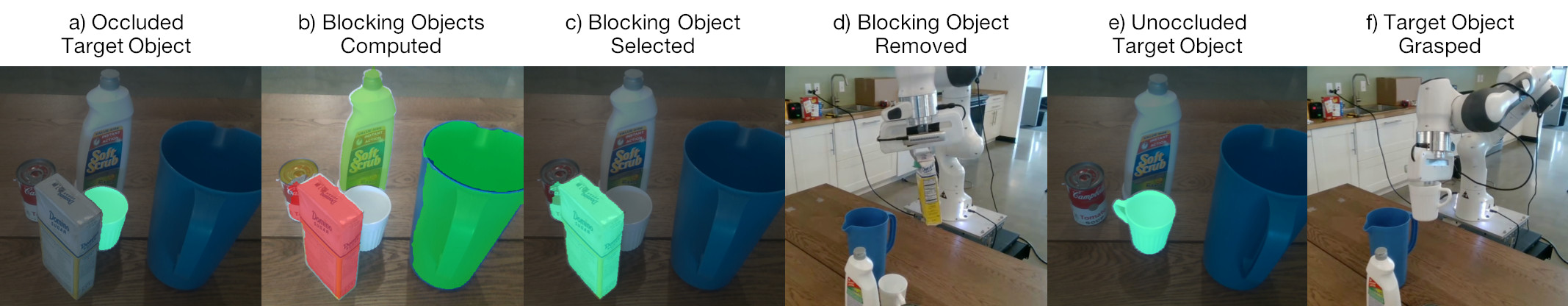}
  \caption{Application of our approach in retrieving a partly occluded mug (highlighted in (a)). The blocking objects are ranked (colored in (b), red being most inhibiting) and removed from the scene. The target object is finally grasped in (f).}
\label{fig:blocker}
\end{figure*}

\begin{figure}
    \centering
    \vspace{-2mm}
    \begin{tabular}{c@{\hskip3pt}c@{\hskip3pt}c}
         \includegraphics[width=0.15\textwidth]{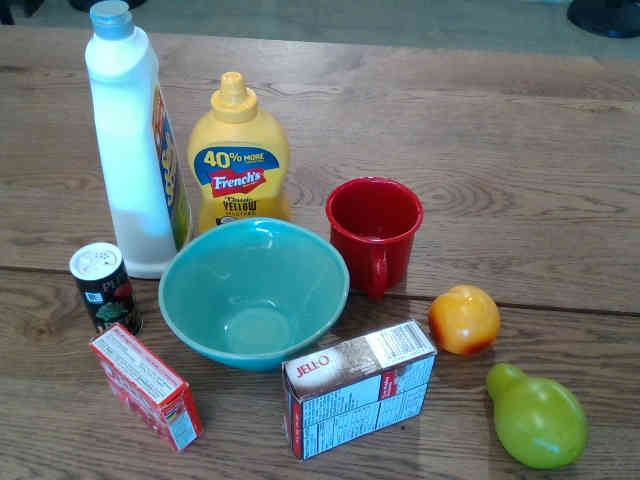}&
         \includegraphics[width=0.15\textwidth]{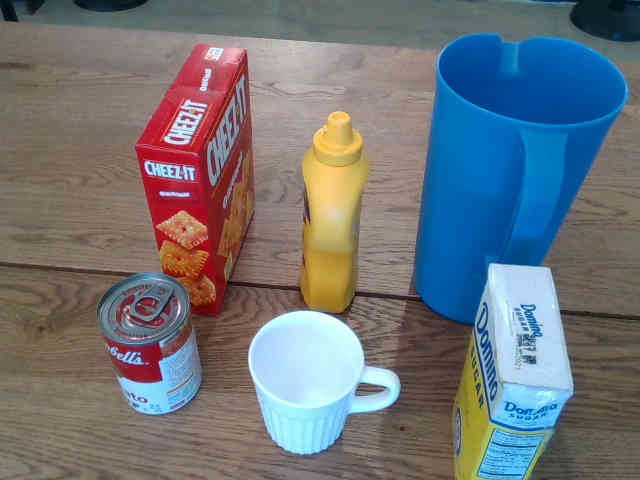}&
         \includegraphics[width=0.15\textwidth]{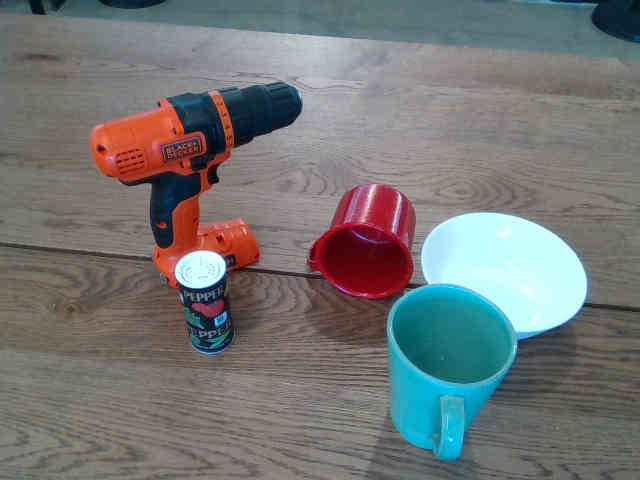}\\
         \includegraphics[width=0.15\textwidth]{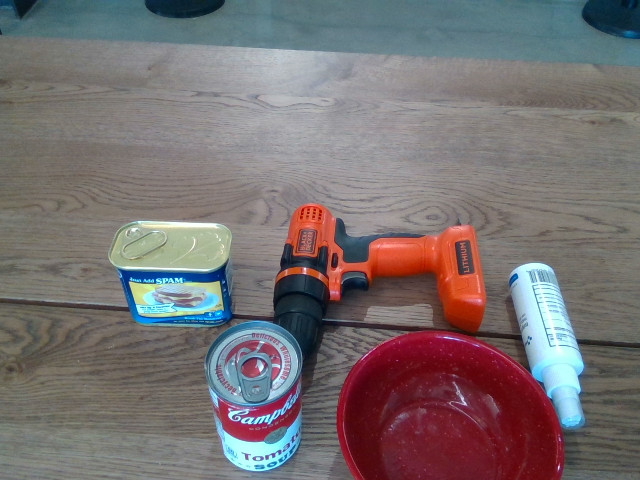}&
         \includegraphics[width=0.15\textwidth]{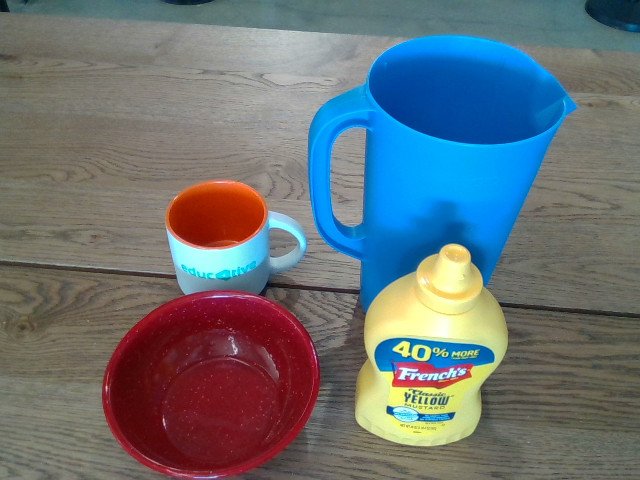}&
         \includegraphics[width=0.15\textwidth]{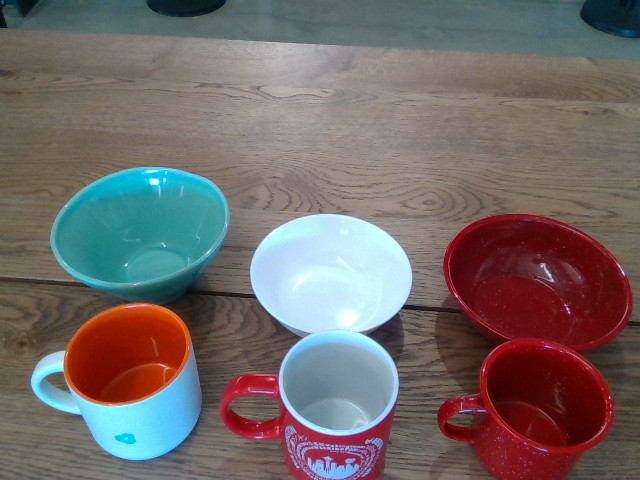}\\
         \includegraphics[width=0.15\textwidth]{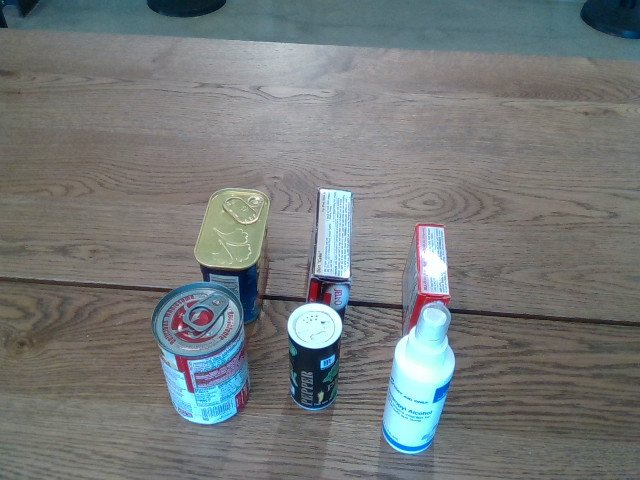}&
         \includegraphics[width=0.15\textwidth]{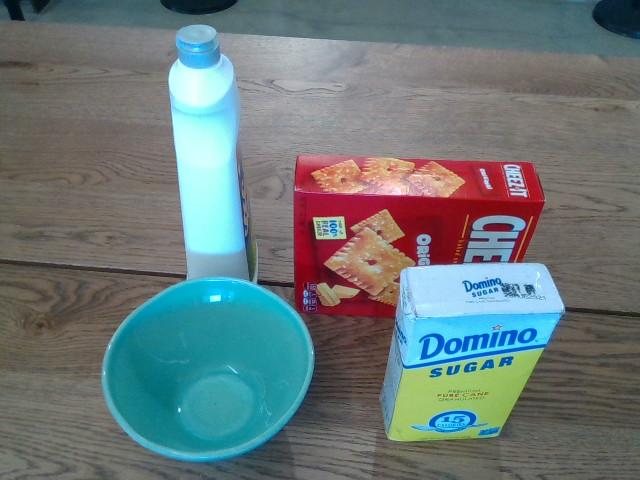}&
         \includegraphics[width=0.15\textwidth]{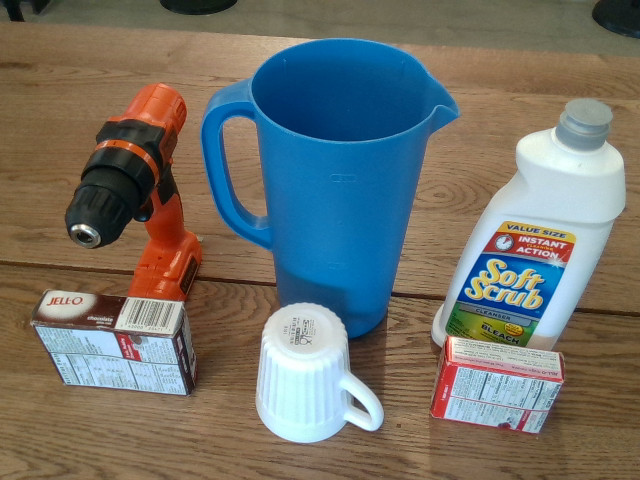}\\
         & 
    \end{tabular}
    \vspace{-3mm}
    \caption{Scenes used for testing. See accompanying video for grasp performance.}
    \label{fig:scenes}
    \vspace{-5mm}
\end{figure}

% \begin{figure}
%   \begin{center}
%     \includegraphics[width = 0.99\linewidth]{figures/testscenes.jpg}
%   \end{center}
%   \caption{Scenes used for testing. See accompanying video for grasp performance. \vspace{-4mm}}
% \label{fig:scenes}
% \end{figure}

\textbf{Role of Object Instance Segmentation:} 
%Target-driven grasping in clutter has two, sometimes competing, objectives: generating robust grasps for the target object while still avoiding collisions with the environment. The former goal only needs instance-level information (such as a segmented point cloud of the object) but the latter requires information of the full scene. The input data used has implications for the learning dynamics of the variational grasp sampler. We trained all the samplers with the same latent size of 2 and there is a lot of pressure during training to map different parts of the latent space to the space of grasps. Using a larger latent space could potentially deteriorate the quality of the sampler grasps during inference as one may end up sampling latent values not seen during training \cite{6dofgraspnet}. 
We compared our cascaded grasp sampling approach to an instance-agnostic baseline. Without instance information, the baseline is a single-stage grasp planner that uses the point cloud of the scene, since we cannot get a object-centric input. An example of the input data to this baseline is shown in Fig.~\ref{fig:segmodes}(a). From the ablation shown in Fig.~\ref{fig:cascaded_single_stage}, we found that our cascaded grasp sampler (using instance information and CollisionNet) had a AUC of 0.22 and outperformed the object instance-agnostic baseline in terms of both success and coverage, which had a AUC of 0.02. A common failure mode of the instance-agnostic model is that the variational sampler gets confused as to which object to grasp in the scene, with the latent space being potentially mapped to grasps for multiple objects and degrading the overall grasp quality for all the objects.

%\textbf{Object-centric vs. Clutter-centric Sampling}: While we want the sampler to generate a diverse set of grasps in 3D (to maximize the set of filtered grasps after evaluation and collision checking), a lot of the grasps sampled from a object-centric model could result in collisions with the environment (e.g. the floor) and clutter. Our sampler is trained with all positive grasps $G_{}^{+}$. An alternative would be to train a clutter-aware sampler that learns to sample collision-free grasps, which will only see positive grasps that do not collide with the environment i.e. $\{g: g \in G_{}^{+}, \neg\Psi(g, \textbf{x})\}$. Fig \ref{fig:curve_vaeconfig} compares the performance of both samplers. While the latter has a slightly higher coverage, it has lower maximum precision and is empirically observed to mainly generate planar grasps. One reason could be VAE does not have enough information of surrounding clutter in object point cloud to make predictions and experiences some form of mode collapse. \adithya{TODO: Report a quantitative number for planar-ness}.

%\begin{figure}
%  \begin{center}
%    \includegraphics[width = 0.8\linewidth]{figures/curve_vaeconfig.png}
%  \end{center}
%  \caption{Success-Coverage curve comparing object-centric and 
%clutter-centric samplers.}
%\label{fig:curve_vaeconfig}
%\end{figure}

\subsection{Real Robot Experiments}
\label{subsec:realrobot}
In our robot experiments, we wanted to show that our cascaded grasp synthesis approach (1) transfers to the real world despite being trained only in simulation; (2) has competitive performance for target-driven grasping in real clutter scenes and (3) outperforms baseline methods using the clutter-agnostic 6-DOF GraspNet implementation \cite{6dofgraspnet} with instance segmentation and voxel-based collision checking. Our physical setup consists of a 7-DOF Franka Panda robot with a parallel-jaw gripper. We used a Intel Realsense D415 RGB-D camera mounted on the gripper wrist for perception. We execute the grasps in a open-loop fashion where the robot observes the scene once, generates the grasps and then executes solely based on the accurate kinematics of the robot. We found open-loop execution to work reasonably well in our setting. CollisionNet only considers collisions between the gripper and the clutter. We also use occupancy voxel-grid collision checking on top of CollisionNet to make sure that the rest of the manipulator arm does not collide with the clutter and table during motion planning.
%\add{While we could have moved the arm around with the eye-in-hand camera to acquire multi-view data or perform 3d reconstruction, such exploratory actions are kinematically infeasible in confined settings. As such, we limit our setup to using only a single-view observation for inference.} We selected a set of \add{23} common household and kitchen objects with varying geometry, physical and visual properties, including some from the YCB dataset \cite{ycb2017}.
We compared the performance of the method on 9 different scenes (see Fig~\ref{fig:scenes}) with the fixed order (pre-computed randomly) of objects to be grasped. A grasp was considered a success if the robot could grasp the object within two consecutive attempts on the same scene. One could choose the order in which all the target objects are completely visible. To make the problem more challenging, half of the chosen target objects were occluded. % in the scenes.
To generate grasps, a batch size of 200 latents were sampled and the grasps that have scores lower than a threshold for each of the evaluator is filtered out. From the remainder of grasps, the one with maximum score is chosen to be executed.
%For all our experiments, we used instance segmentation from \cite{chrisSegmentation}. We also tested our cascaded approach without using CollisionNet to further emphasize the effectiveness of the collision evaluator. To test target-driven grasping, we specifically curate different cluttered scenes and gave a random priority ordering for grasping the objects. Examples of scene can be scene in Fig.~\ref{fig:scenes}.
%It is noteworthy that this setting is more challenging than just trying to grasp the closest object, which would typically be unoccluded and reducing the complexity of perception and collision avoidance.
%We evaluated for a total of 51 picks in 9 cluttered scenes. For the three different models we tested, we ensured that the initial poses of the objects are approximately the same for each clutter scene. 
%For testing, w
\begin{table}[H]
\footnotesize
\centering
\caption{Real Robot experiments}
% \vspace{-0.5em}
\label{tb:real}
\scalebox{0.9}{
\begin{tabular}{ccc}
\toprule
\textbf{Approach} &  \textbf{Trials}  & \textbf{Performance (\%)} \\
\hline
6-DOF GraspNet \cite{6dofgraspnet} + Ins. Segmentation \cite{chrisSegmentation}                & 32/51       & 62.7    \\
Object Instance              & 31/51       & 60.7\\  
Object Instance + CollisionNet (Ours)                & 41/51 & 80.3       \\
\bottomrule
\end{tabular}}
% \vspace{-1.5em}
\end{table}

The results are summarized in Table~\ref{tb:real}. Our framework achieves a success rate of 80.3$\%$ and outperforms the baseline 6 DOF-GraspNet approach by 17.6$\%$. Furthermore, without CollisionNet, our model performance degrades substantially. The two failure cases are the grasps that are colliding with the object but object centric evaluator predicts high score for them. These grasps are filtered out by CollisionNet. The second failure mode pertains to the fact that the voxel-based representation cannot capture all collisions.

\subsection{Application: Removing Blocking Objects}
\label{subsec:blocker}
%Consider the scenario, such as in Fig \ref{fig:blocker}, 
Consider scenarios such as that shown in Fig.~\ref{fig:blocker},
where the target object is being blocked by other objects and none of the sampled grasps are kinematically feasible. To accomplish this task, the model needs to generate potential grasps for the target object even though the target object is not physically reachable (detected by low scores from CollisionNet). Given the potential grasps, we can identify which objects are interfering with the generated grasps for the target object. The blocking object~$j$ is chosen to be the one with the largest increase in collision scores when removing the corresponding object points from the scene point cloud i.e. $\alpha_{j} = P(C|\hat X_{j}, g) - P(C|X, g)$. The objects are colored by this ranking metric $\alpha_{j}$ in Fig.~\ref{fig:blocker}(b), with the red object being the most blocking. The modified point cloud $\hat X_{j}$, which hallucinates the scene without object~$j$, is implemented by merging the object's instance mask with that of the table and projecting corresponding points to the table plane. Grasps are then generated for the blocking object and removed from the scene. Collision-free grasps can now be generated for the unoccluded target object for the robot to recover it. Target objects can be specified by any down-stream task but in this use case, it is specified by choosing the segmentation corresponding to the target object.

\section{Conclusion}

We present a learning-based framework for 6-DOF grasp synthesis for novel objects in structured clutter settings. This problem is especially challenging due to occlusions and collision avoidance which is critical. Our approach achieves a grasp accuracy of 80.3$\%$ in grasping novel objects in clutter on a real robotic platform despite being only trained in simulation. A key failure mode of our approach is that it only considers gripper pre-shape collisions by design and hence motion planning could still fail on generated grasps. In future work, we hope to consider trajectory generation in grasp planning and explore the use of our approach in task planning applications. We also aim to apply this framework in grasping objects from challenging environments like kitchen cabinets and handle the case of retrieving stacked objects in structured clutter.

% \item A learning-based approach for 6-DOF grasp synthesis for novel objects in structured clutter.
% \item Propose a learnt collision checker, conditioned on the gripper information and on the raw partially-occluded point cloud of the scene
% \item Show that our approach, trained only with synthetic data, achieves a grasp accuracy of 80.3$\%$ with 24 real-world novel objects in clutter. It also outperforms a clutter-agnostic baseline approach of 6-DOF GraspNet \cite{6dofgraspnet} with state-of-the-art instance segmentation \cite{chrisSegmentation} by 17.6\%.
% \item Demonstrate an application of our approach in moving blocking objects away out of the way to grasp a target object that is initially occluded and impossible to grasp.

% ~\newline
% \small
% \noindent
% \textbf{ACKNOWLEDGEMENTS:} The authors would like to thank Ankur Handa, Tucker Hermans, Xinke Deng, Christopher Xie and Jonathan Tremblay for discussions and feedback.

%===============================================================================

\bibliographystyle{IEEEtran}
\bibliography{IEEEabrv,references}

%===============================================================================

\end{document}